\documentclass[letterpaper, 10 pt, conference]{ieeeconf}  

\IEEEoverridecommandlockouts                              

\usepackage{graphicx}
\graphicspath{ {./images/},{./images/gta2cs/}}

\usepackage{epsfig} 
\usepackage{mathptmx} 
\usepackage{times} 
\usepackage{amsmath} 
\usepackage{amssymb}  
\usepackage{caption}
\usepackage{float}

\newcommand{\norm}[1]{\left\lVert#1\right\rVert}

\title{\LARGE \bf
Adversarial Appearance Learning in Augmented Cityscapes for Pedestrian Recognition in Autonomous Driving
}

\author{Artem Savkin$^{1,2}$ and Thomas Lapotre$^{1}$ and Kevin Strauss$^{1}$ and Uzair Akbar$^{1}$ and Federico Tombari$^{1,3}$
	\thanks{$^{1}$TU Munich, Boltzmannstr. 3, 85748 Munich (Germany)
		{\tt\small artem.savkin@tum.de},
		{\tt\small thomas.lapotre@tum.de},
		{\tt\small uzair.akbar@tum.de},
		{\tt\small kevin.strauss@tum.de},
		{\tt\small tombari@in.tum.de}}%
	\thanks{$^{2}$BMW AG, Munich (Germany)}
	\thanks{$^{3}$Google, Zurich (Switzerland)}
}

\begin{document}
\maketitle

\begin{abstract}
In the autonomous driving area synthetic data is crucial for cover specific traffic scenarios which autonomous vehicle must handle. This data commonly introduces domain gap between synthetic and real domains. In this paper we deploy data augmentation to generate custom traffic scenarios with VRUs in order to improve pedestrian recognition. We provide a pipeline for augmentation of the Cityscapes dataset with virtual pedestrians. In order to improve augmentation realism of the pipeline we reveal a novel generative network architecture for adversarial learning of the data-set lighting conditions. We also evaluate our approach on the tasks of semantic and instance segmentation.

\end{abstract}

\section{Introduction}

Transiting from the mere research area to application domain autonomous driving must be able to handle broad variety of traffic scenarios occurring in the real world. This represents one of the main challenges in developing the autonomous vehicles and theirs perception components. The ability to master yet unseen situations will determine if such a self-driving vehicle is able to navigate fully autonomously.

Nowadays neural networks have largely improved generalization capabilities of perception systems of autonomous vehicles. Their performance and generalization capacity is, however, known to be very data reliant. This property highly influences development and quality assurance of perception algorithms. In order to ensure that a particular scenario will be handled by a perception model such scenario should be available in the training dataset beforehand. Engineers constantly improve their models by capturing new scenarios on the street and integrating them into the model via fine-tuning. This approach involves annotation of any newly captured scene every single time.

Although annotation procedure seems to be quite straight-forward it is not always the case in the real world. Labelling becomes a non-trivial task when it comes to "long-tail" scenarios which occur rarely and are hard to capture. Another challenging task is to cover near-accident scenarios which could not be replicated due to ethical reasons, e.g. endangering vulnerable traffic users (VRUs).

For previously described examples synthetically generated data seem to be very promising solution, as any rare or near-accident scenario could be simulated. Many researchers utilized the idea of simulated data in theirs works: \cite{Barron1992} \cite{Broggi2005}, \cite{Taylor2007}, \cite{Liebelt2008}. However appealing this approach might seem at the first glance it hardly find its application in the real world. The main reason for that is so called \textit{domain shift} introduced by simulation. Multiple works have shown that the models trained merely on synthetic data show drastic performance drop compared to the ones trained on real data when evaluated also on real data. \cite{Richter2016} reports almost 20\% \textit{meanIoU} decrease for semantic segmentation model \cite{Kundu2016} when it has been trained on CamVid(train) \cite{Brostow2009} and synthetic dataset \cite{Richter2016} but evaluated on CamVid(val). Simulated data is typically different from real one with regard to content distribution and appearance. This phenomenon is commonly addressed as \textit{covariate shift} and it is considered to be the main reason for such performance drop \cite{Dundar2018}.

\begin{figure}
\begin{minipage}{0.495\columnwidth}
\includegraphics[width=1\textwidth]{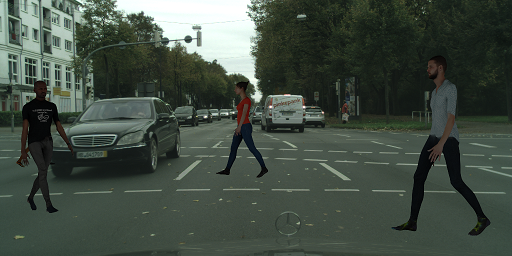}
\caption*{Augmented Image}
\includegraphics[width=1\textwidth]{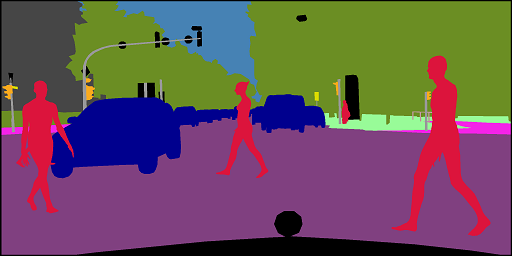}
\caption*{Semantic Ground Truth}
\end{minipage}
\begin{minipage}{0.495\columnwidth}
\includegraphics[width=1\textwidth]{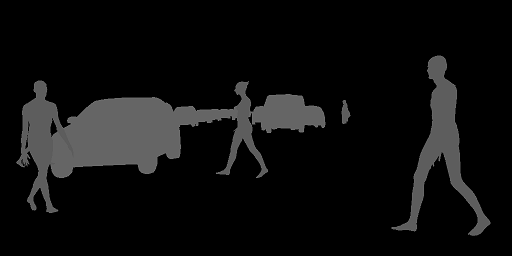}
\caption*{Instance Ground Truth}
\includegraphics[width=1\textwidth]{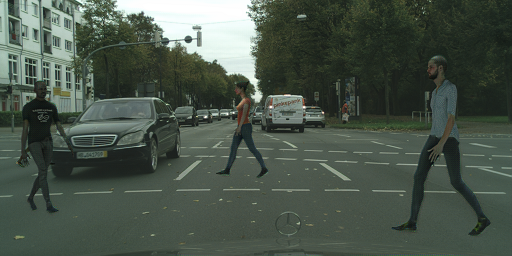}
\caption*{Translated Image}
\end{minipage}
\caption{Example of augmented image together with semantic and instance ground truths and original image after translation}
\label{fig:teaser}
\end{figure}

\begin{figure}[t!]
\begin{minipage}{0.325\columnwidth}
\includegraphics[width=1\textwidth]{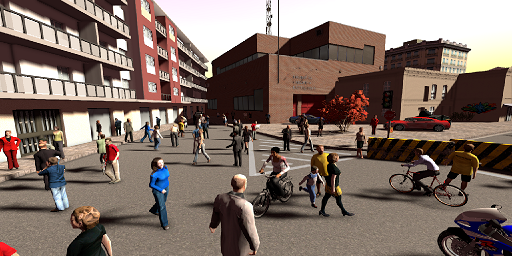}
\includegraphics[width=1\textwidth]{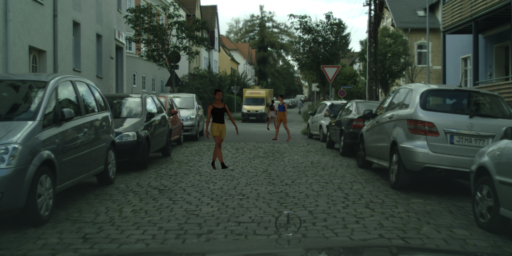}
\includegraphics[width=1\textwidth]{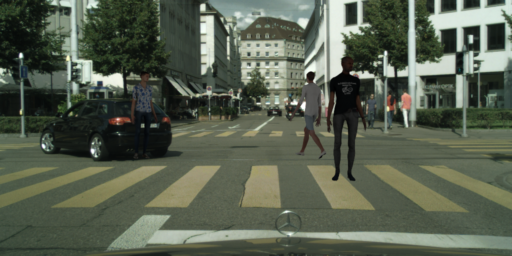}
\caption*{Original}
\end{minipage}
\begin{minipage}{0.325\columnwidth}
\includegraphics[width=.5\textwidth]{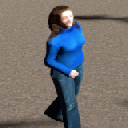}\includegraphics[width=.5\textwidth]{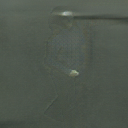}
\includegraphics[width=.5\textwidth]{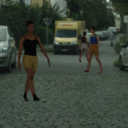}\includegraphics[width=.5\textwidth]{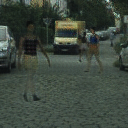}
\includegraphics[width=.5\textwidth]{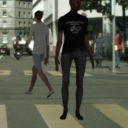}\includegraphics[width=.5\textwidth]{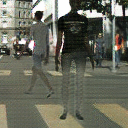}
\caption*{Zoom In}
\end{minipage}
\begin{minipage}{0.325\columnwidth}
\includegraphics[width=1\textwidth]{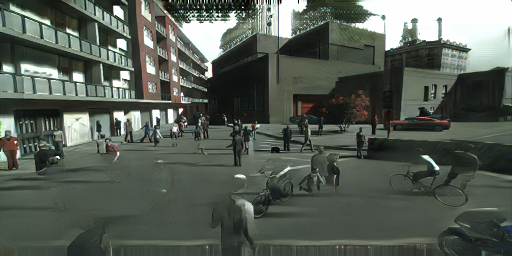}
\includegraphics[width=1\textwidth]{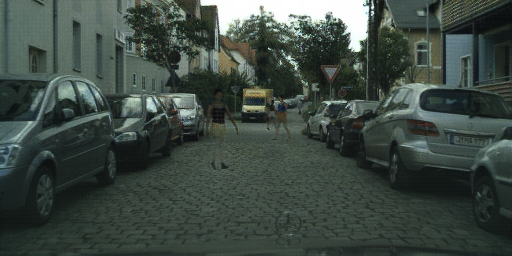}
\includegraphics[width=1\textwidth]{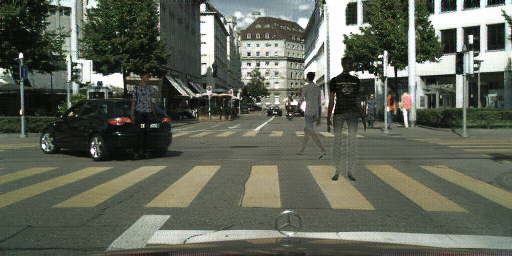}
\caption*{Generated}
\end{minipage}
\caption{Examples of semantic inconsistencies introduced by adversarial training.}
\label{fig:vanishing}
\end{figure}

In order to tackle the aforementioned problem many researchers attempt to minimize the gap between synthetic and real domains by means of generative networks. The majority of them rely on adversarial training \cite{Goodfellow2014}, \cite{Shrivastava2016}, \cite{Zhu2017}, which is, however, a subject for \textit{covariate shift} itself. Although adversarial nets learn to deliver visually realistic data, such models suffer from semantic mismatch and tend to integrate visual artifacts into generated samples. Several examples of the mentioned image perturbations are shown on the figure~\ref{fig:vanishing}. Thus, multiple works are focused on improving adversarial networks by integrating certain constraints in order to level introduced inconsistencies \cite{Hoffman2017}.

In this work we focus on pedestrian detection and provide an augmentation pipeline which allows to enhance real dataset such as Cityscapes \cite{Cordts2016} with virtual pedestrians in different scenarios. This pipeline makes geometrically correct in-painting of the pedestrian CAD models into Cityscapes scenes possible. Yet augmentation does not consider lighting conditions of the particular dataset scene. Hence, as a second contribution of our paper we provide a domain adaptation model which is based on an adversarial network and allows to learn appearance and make CAD model realistic. Our adversarial network employes multiple discriminators \cite{Karras2018ICLR} such as used for multiple image resolutions. But due to introduced masking approach our model is robust to distribution discrepancies between the real and synthetic datasets and it is able to produce consistent imagery with realistic lighting conditions and appearance of the original dataset. An example of an augmented and adapted samples alongside with corresponding semantic and instance segmentation ground truths are shown in the figure~\ref{fig:teaser}. We claim that our approach allows to simulate various scenarios with vulnerable road users without introducing simulation-to-real gap. Our evaluation section demonstrates this in details.

\begin{figure*}[t]
\begin{minipage}{0.33\textwidth}
\includegraphics[width=1\textwidth]{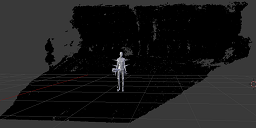}
\caption*{Point Cloud (black)}
\end{minipage}
\begin{minipage}{0.33\textwidth}
\includegraphics[width=1\textwidth]{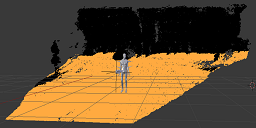}
\caption*{Spawn Map (orange)}
\end{minipage}
\begin{minipage}{0.33\textwidth}
\includegraphics[width=1\textwidth]{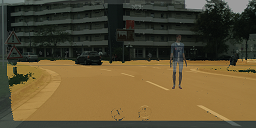}
\caption*{Blending}
\end{minipage}
\caption{Visualization of 3D model placement pipe-line with reconstructed spatial information about the scene: stereo point cloud (black), estimated collision free \textit{spawn map} (orange), and render of 3D model blended with the scene frame.}
\label{fig:augmentation}
\end{figure*}
\section{Related Work}
Application of synthetic data in autonomous driving domain and beyond has already a comparably long history.

Recent research CAD models gained a lot of atention. In order to accurately predict human 3D pose \cite{Shotton2013} used synthetic training set of high variability to learn models invariant to poses, view and other factors. \cite{Varol2017} provided the whole dataset based on realistic human augmentations called SURREAL. Unlike 3D pose estimation \cite{Aubry2014} and \cite{Sun2014} used 3D models for multi-category 2D object detection.

Due to extreme complexity of the data and vast variety of possible scenarios the idea of synthetic data generation was extremely appealing in the domain of autonomous driving. It has been utilizing virtual simulators, such as \cite{Chen2015} and CARLA \cite{Dosovitskiy2017}. Recently multiple works revealed synthetic datasets completely consisting of rendered imagery. One of them is SYNTHIA \cite{Ros2016} which provided 9400 frames of scenes in different lighting and weather conditions. Another dataset \cite{Richter2016} utilized video game engine to annotate 25000 images and \cite{Gaidon2016} simulated a clone of real KITTI dataset \cite{Geiger2012}.

Some works simplified their task and picked rendering single objects instead of rendering the whole image. So \cite{Hattori2015} used simulated pedestrians to learn pedestrian detector for a surveillance system. Another usage of virtual humans has been revealed in \cite{Chen2016} aiming at human pose estimation. MixedPeds from \cite{Cheung2017} exploited the idea pedestrian augmentation in the domain of autonomous driving. \cite{Huang2017} specialized on synthetic pedestrians in unexpected traffic situations while enhancing real data and \cite{Alhaija2017} has been using car augmentation for car instance segmentation. Spatial and semantic correctness for inserting objects into the image has been approached by an end-to-end trainable network from \cite{Lee2018}.

The aforementioned approaches produced very useful training data, yet still distinguishable from actual real world imagery. More sophisticated approaches tried to learn the appearance and synthesize already target-alike imagery. Vast majority of them are based on generative models and leverage adversarial network architecture. So \cite{Alhaija2018} introduced augmentation network which synthesizes geometrically consistent yet realistic in-paintings of cars. Other works focused on generating the whole image to be target data alike. For example, Pix2Pix in \cite{Liu2017} introduced conditioning of an adversarial training on the prior information about semantic layout. Another conditioning strategy has been proposed in \cite{Qi2018}, where they used predefined patches from database to create image canvas. \cite{Wang2017} has improved Conditional GANs and integrated instance information together with image manipulation possibility. \cite{Zhu2017} added cycle loss additionally to adversarial one to achieve stable and consistent image transfer results. Further works like \cite{Liu2017}, \cite{Shu2018}, based on the tandem of cycle loss and adversarial loss tried to disentangle appearance and content by learning the latent representation space.

\section{Approach}

Our data generation pipeline consists of 2 parts - data augmentation phase and appearance learning phase.

In the first phase we set virtual VRUs (vulnerable road users) into the existing scene and blend the resulting rendering with the camera snapshot of the scene (actual frame). This part focuses on geometrical and optical correctness of the blending process. Virtual pedestrians are intended to be allocated only to appropriate locations in the scene such as sidewalks or roads. Placement of the 3D models also requires avoiding collisions among them as well as with objects already existing in the scene (cars, trees, poles, etc.) To achieve optical correctness of the augmentation process a rendering camera is calibrated with the parameters of the dataset camera. This phase relies only on spatial information about the scene and disregards any appearance characteristics such as lighting conditions. The results of this phase are real dataset frames with virtual pedestrians on it which are clearly distinguishable by their synthetic optics.

In the second phase we aim to learn visual features of the dataset scenes and apply them onto the in-painted pedestrians so that they can attain a realistic look. This part is based on a dedicated adversarial network architecture which is considered to be robust to distribution discrepancy between synthetic and real domains. For our adversarial architecture we adopt a generator network from \cite{Zhu2017} which showed very stable and visually appealing results, and for the discriminator part we deploy our multi-discriminator architecture which will be described in the corresponding subsection~\ref{sec:domain_adaptation}.

\subsection{Data Augmentation}

\subsubsection{Spawn Map}
Dataset augmentation phase starts with the estimation of a so called \textit{spawn map}. The main goal of this stage is to compute suitable spots where virtual pedestrian could be located without the risk of a collision with other objects which do already exist in the scene such as buildings, vegetation or cars. In order to estimate spawn map one relies on spatial information about the scene. This information could be typically directly obtained either from a LIDAR or a stereo camera. In our approach we leverage spatial data reconstructed from disparity maps obtained from stereo cameras from Cityscapes dataset.

Basing on the disparity maps and camera extrinsic and intrinsic parameters we calculate corresponding depth map for every particular image. An example for a point cloud reconstructed from such depth map can be observed in figure~\ref{fig:augmentation}.

Utilizing simple threshold-based heuristic together with isolation forest algorithm we estimate ground level and eliminate outliers. Ground level including sidewalk and road surface allows us to put pedestrians without the risk of being located onto unsuitable spots such as buildings or walls. Having \textit{the spawn map} at our disposal we now can sample any location to put a pedestrian semantically correct at said location.

\subsubsection{Collision Tracking}
In addition, to that we aim to avoid overlapping among virtual pedestrians as well as overlapping with other dynamic objects in the scene. Any location in the scene which is already taken is tracked using a \textit{collision map}. This builds some kind of a \textit{free space} representation. Initially any reconstructed stereo point is put into the collision map if it does not belong to the spawn map. Any pedestrian placed into the scene will extend the \textit{collision map} in order to prevent intersection with new meshes.

\subsubsection{Blending}
In the final step the layer with the 3D pedestrian model will be rendered and blended with existing frame capture from dataset. This together with the described pedestrian placement process involving point cloud reconstruction, spawn map and collision map estimation and located virtual pedestrians is visualized in figure~\ref{fig:augmentation}. It is necessary to notice here, that in order to achieve optical correctness while blending, the rendering camera shall be configured with the extrinsic and intrinsic parameters of the original camera which has captured the frame.

\subsection{Domain Adaptation}\label{sec:domain_adaptation}
In the domain adaptation part of our pipeline, we intend to learn appearance characteristics of the target dataset and apply them to the augmented models. Here pedestrians should gain a realistic look and realistic lighting.

\subsubsection{Vanishing Pedestrians}
Adaptation part is based on the widely used adversarial training \cite{Goodfellow2014}, which involves 2 neural networks acting against each other. The first one is called generator and it gets samples from source data or noise vectors to synthesize an image indistinguishable from the target data. The second one, called discriminator, in turn tries to tell apart synthesized and target images and penalize generator. During the training it aims to find equilibrium in a zero-sum game between networks and continues until in ideal case discriminators prediction is equivalent to random guessing.

When adversarial training converges it provides a set of synthesized samples with distribution similar to the target data. Although adversarial training is a powerful tool in a domain adaptation task it commonly creates synthetic mismatches between source and generated data. Whilst the generator is encouraged to perform perturbations in the source images, the discriminator identifies discrepancies between 2 domains very well and guides those perturbations to level out the discrepancies.

Such behavior can be observed in multiple domain adaptation setups. Examples of adversarially introduced inconsistencies in synthetic to real domain adaptation are shown in figure~\ref{fig:vanishing}. In general this can heavily impact learning the procedure with synthesized data as source ground truth does not match corresponding images any more. In our specific augmented-to-real adaptation setup the discriminator easily detects foreign pedestrians and encourages the generator to render them away in order to restore initial distribution. This has a rather undesired effect in our data generation pipeline, since the in-painted objects as shown in figure~\ref{fig:vanishing} vanish.

\begin{figure}[t]
\includegraphics[width=1\columnwidth]{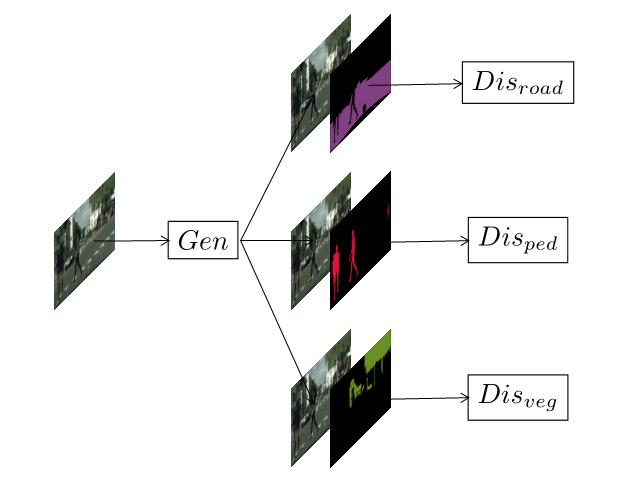}
\caption{Multi-discriminator architecture with generator (Gen) and 3 class specific discriminators (Dis).}
\label{fig:architecture}
\end{figure}

\subsubsection{Multi-Discriminator}\label{sec:multi_discriminator}
In our approach we propose a novel GAN model which is designed to handle distribution discrepancies by splitting the discriminator into multiple class-specific ones. The intuition behind this method is to reduce the decision-making freedom of the discriminator to overcome distribution discrepancy between source and target datasets. We suggest eliminating the degree of freedom with regard to content and let the discriminator only focus on the appearance features of a particular classes.

For that matter we extend the number of discriminators but at the same time we let each of them only assess one part of the generated image which corresponds to a particular semantic class. This could be achieved by splitting the synthesized image into multiple disjoint patches, where every patch contains one semantic class and providing only that particular patch to a corresponding discriminator $D_r^c$. The full view of the multi-discriminator architecture is depicted in figure~\ref{fig:architecture}.

In this illustration one can see a generator $G_r$ which is provided with the image from our augmented dataset as an input. The generator translates augmented images into real ones and then its output will be split into multiple patches using semantic maps. Each one is provided to a dedicated discriminator $D_r^c$ after that. Training such a multi-discriminator $D_r$ makes each split discriminator $D_r^c$ get specialized on the appearance features of one given class. To obtain this we optimize a special aggregated objective.

Our aggregated objective $\mathcal{L}_{adv}$ provided for the particular multi-discriminator is composed from all class-specific objectives:

\begin{equation}
\mathcal{L}_{adv}(G_r, D_r) = \sum_{c}^{N_c} \mathcal{L}(D_r^c,G_r)
\label{eq:objective}
\end{equation}

In \ref{eq:objective}, $c$ represents a particular class and $N_c$ is an overall number of classes in a dataset. In our task of transferring augmented images, the number of classes $c = 2$, so we simplify the architecture to only 2 discriminators: one for the pedestrian class and one for the rest. The simplified version of objective~\ref{eq:objective} will be accordingly reduced to:

\begin{equation}
\mathcal{L}_{adv}(G_r, D_r) = \mathcal{L}(D_r^p,G_r) + \mathcal{L}(D_r^r,G_r)
\label{eq:aug_objective}
\end{equation}

\subsubsection{Masking}
\begin{figure}[b!]
\includegraphics[width=1\columnwidth]{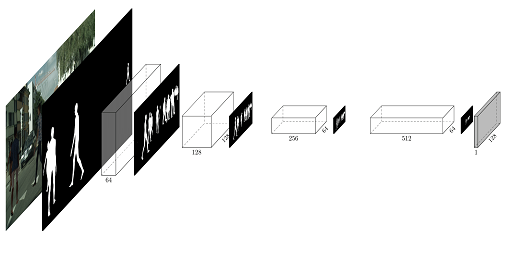}
\caption{Overview of the class specific discriminator with its \textit{MaskLayer} introduced after each convolution block.}
\label{fig:masked_discriminator}
\end{figure}

For every class specific discriminator we adopt the PatchGAN architecture from \cite{Isola2016}. To ensure that each of them is provided with only single-class patches of the input image we mask out all pixels of irrelevant classes by replacing them with 0 values. Masking only input images, however, does not reduce propagation of the signal from the whole image. Hence, we clip activations that originate from undesired regions of the input image at each level of the discriminator. As to achieve that we introduce a \textit{MaskLayer} and apply a down-sampled version of the original mask $M_c$ with $c$ denoting a particular class on the feature maps after each convolution layer of discriminator. More detailed overview of a class-specific discriminator architecture is shown in figure~\ref{fig:masked_discriminator}.

\subsubsection{Cost-sensitive loss}
Our adaptation pipeline consists of pairs of input images $x^i$ of size $3 \times h \times w$ together with corresponding labels $y^i$ from augmented dataset: $\{(x_a^i, y_a^i)\}_{i=1}^{N_a}$ and pairs $\{(x_r^j, y_r^j)\}_{j=1}^{N_r}$ from real domain.

Random variable $x$ take values $x_a^i$ in the input distribution space $\mathcal{X}$ and $y_a^i$ in labels distribution spaces $\mathcal{Y}$, which are \textit{independent and identically distributed} and follows joint probability distribution $P_a(x,y)$:

\begin{equation}
\begin{split}
&x_a^i \in \mathcal{X}_a \subset \mathcal{X} \subset \mathbb{N}^{3\times h \times w}, i = 0,1, ..., N_a\\
&y_a^i \in \mathcal{Y}_a \subset \mathcal{Y} \subset \mathbb{N}^{h \times w}, i = 0,1, ..., N_a\\
&\{x_a^i, y_a^i\}_{i=1}^{N_a} \sim P_a(x,y)
\end{split}
\end{equation}

The real samples $x_r^j$ in turn follow different probability distribution $P_r$:

\begin{equation}
\begin{split}
&x_r^j \in \mathcal{X}_r \subset \mathcal{X} \subset \mathbb{N}^{3\times h \times w}, j = 0,1, ..., N_r\\
&y_r^j \in \mathcal{Y}_r \subset \mathcal{Y} \subset \mathbb{N}^{h \times w}, j = 0,1, ..., N_r\\
&\{x_r^j, y_r^j\}_{j=1}^{N_r} \sim P_r(x,y)
\end{split}
\end{equation}

In our class-specific discriminator we compute the error values with regard to only relevant regions, so we apply masks $M_c$ to error calculation as well. Hence, class discriminator objective looks as follows:

\begin{equation}
\begin{split}
&\mathcal{L}(D_r^c,G_r) =\\
&\mathbb{E}_{(x_r, y_r)} \left[ \frac{1}{wh} \lVert D_r^c(x_r, M^c) \circ M^c(y_r) \lVert_{F^2} \right] +\\
&\mathbb{E}_{(x_a, y_a)} \left[ \frac{1}{wh} \lVert (D_r^c(G_r(x_a), M^c)-J) \circ M^c(y_a) \lVert_{F^2} \right]\\
\end{split}
\end{equation}\\
Here, $J$ denotes a ones matrix of size $h \times w$ and $\lVert \cdot \lVert_{F^2}$ is a Frobenius norm. We intentionally keep \textit{masked MSE} normalized by the size of the actual sample, since this leads to the fact that masks of different size contribute differently to the particular loss. We encourage our model to learn more from samples with more prominent instances of class of interest (e.g. pedestrian), which provides more information about appearance.

A naive application of this objective function in the adversarial training procedure naturally results in putting more emphasis on the background that usually covers the major part of an image (road, building, etc.). In the case of \textit{augmented Cityscapes} dataset, around 95\% of pixels represent non-pedestrians classes. This makes the problem highly unbalanced where the non-pedestrians pixels would contribute 19 times more intense to the overall objective. Thus, we want to eliminate the effect of dominating classes on the dataset scale as we deal with class unbalanced data.

This could be achieved by means of weighting factor $\lambda$ as a hyper-parameter. Our experiments show that it works best if $\lambda$ reflects actual class ratio in the dataset:
\begin{equation}
\label{unbalance}
\lambda = \frac{\sum_y \norm{M_p(y)}_{1}}{\sum_y \norm{M_r(y)}_{1}}
\end{equation}
Analogous calculation is applied in the case of multi-discriminator with more than 2 classes of interest.\\
Thus, the overall \textit{cost-sensitive objective} takes form:
\begin{equation}
\begin{split}
&\mathcal{L} = \lambda_{cyc}\mathcal{L}_{cyc} +\\
&\mathcal{L}_{adv}(D_r^p,G_r) + \lambda \mathcal{L}_{adv}(D_r^r,G_r) +\\
&\mathcal{L}_{adv}(D_a^p,G_a) + \lambda \mathcal{L}_{adv}(D_a^r,G_a) +\\
\label{eq:}
\end{split}
\end{equation}
Here $\mathcal{L}_{cyc}$ represents the \textit{cyclic-consistency loss} together with its weight $\lambda_{cyc}$ introduced by \cite{Zhu2017}. Now when the objective is defined, we follow adversarial training procedure to optimize:
\begin{equation}
\min\limits_{G_r,G_a} \max\limits_{D_r^p, D_a^p, D_r^r, D_a^r} \mathcal{L}(G_r,G_a,D_r^p, D_a^p, D_r^r, D_a^r)
\end{equation}

\begin{table}[!t]
\renewcommand{\arraystretch}{1.5}
\resizebox{\columnwidth}{!}{
\begin{tabular}{l|cc|cc}
\hline
\rotatebox[origin=l]{0}{ Method }
& \rotatebox[origin=c]{0}{ $AP_{avg}$ }
& \rotatebox[origin=c]{0}{ $AP50_{avg}$ }
& \rotatebox[origin=c]{0}{ $AP_{person}$ }
& \rotatebox[origin=c]{0}{ $AP50_{person}$ }
\\
\hline

CS		& 31.8 & 59.0 & 33.0 & 67.7 \\

Ours	& 32.6 & 60.4 & 35.6 & 74.2 \\

\hline
\end{tabular}
}
\caption{Instance segmentation results for Mask-RCNN trained on Cityscapes(top) and our dataset (bottom)}
\label{tab:maskrcnn}
\end{table}

\section{Experiments}

We base our experiments on the popular public dataset Cityscapes \cite{Cordts2016}. It provides all the data necessary for the data generation: camera frames and disparity maps alongside with calibration parameters to enable augmentation as well as computer vision task ground truth to perform evaluation on it.

\subsection{Datasets}
Dataset of our choice must meet several criteria: it must provide information about spatial characteristic of the scene which will be augmented, it also shall provide ground-truth to preform computer vision task and evaluation with generated data.

Cityscapes is a large-scale computer vision dataset which provides 5000 camera snapshots of the size $2048\times1024$ pixels alongside with dense pixel segmentation labels. The annotations provide ground-truth for both semantic and instance segmentation. The dataset enfolds measurements in cities across Germany (and Strasbourg) and during various seasons as well as weather and daytime conditions. Images are split into \textit{train}, \textit{val} and \textit{test} sets with 2975, 500 and 1525 samples respectively. Together with fine annotations the dataset provides 20,000 weakly annotated images, but the latter ones are not used. Semantic maps contain information about 30 classes such as \textit{road, sidewalk, person, car, etc.} and instance segmentation involve 8 categories: \textit{person, rider, car, truck, bus, train, motorcycle, bicycle}.

With the proposed augmentation pipeline we generate \textit{augmented cityscapes} dataset consisting of 2975 images of resolution 2048 $\times$ 1024 where we randomly insert 1 ~ 5 virtual pedestrians. All images are accompanied with generated both semantic and instance maps. We follow standard Cityscapes annotation format regarding the classes and categories.\\
On having generated \textit{augmented cityscapes} we train \textit{multi-discriminator GAN} model, introduced in section~\ref{sec:domain_adaptation}, for 200 epochs on it. The model was trained from scratch with \textit{cyclic weight} set to 10 and \textit{cost-sensitive} $\lambda = 0.2$. Just like in \cite{Zhu2017} we start with the learning rate of 0.0002 and keep it constant for the first 100 epoch, decaying to 0 in the course of another 100. For training we downscale images by a factor of 2 and use no random crops.

\begin{figure}[!t]
\begin{minipage}{0.325\columnwidth}
\includegraphics[width=1\textwidth]{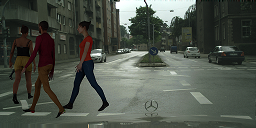}
\includegraphics[width=1\textwidth]{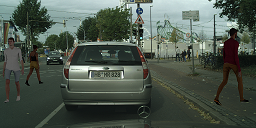}
\includegraphics[width=1\textwidth]{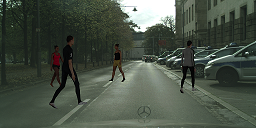}
\includegraphics[width=1\textwidth]{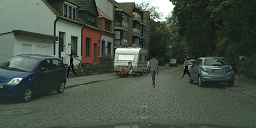}
\caption*{Original Image}
\end{minipage}
\begin{minipage}{0.325\columnwidth}
\includegraphics[width=.5\textwidth]{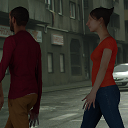}\includegraphics[width=.5\textwidth]{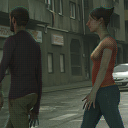}
\includegraphics[width=.5\textwidth]{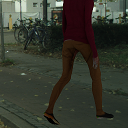}\includegraphics[width=.5\textwidth]{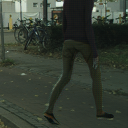}
\includegraphics[width=.5\textwidth]{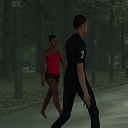}\includegraphics[width=.5\textwidth]{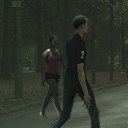}
\includegraphics[width=.5\textwidth]{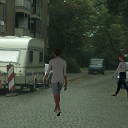}\includegraphics[width=.5\textwidth]{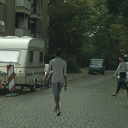}
\caption*{Zoom}
\end{minipage}
\begin{minipage}{0.325\columnwidth}
\includegraphics[width=1\textwidth]{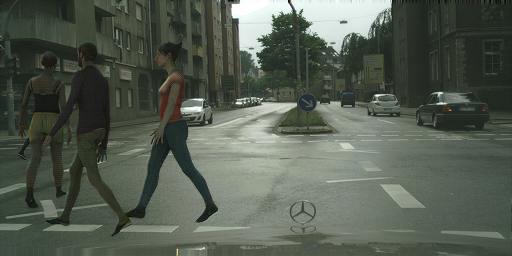}
\includegraphics[width=1\textwidth]{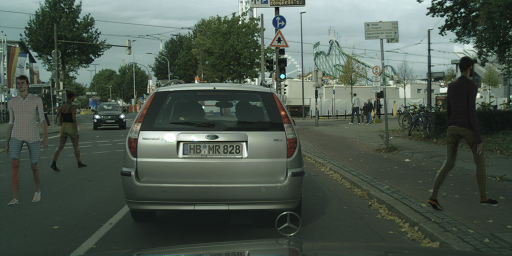}
\includegraphics[width=1\textwidth]{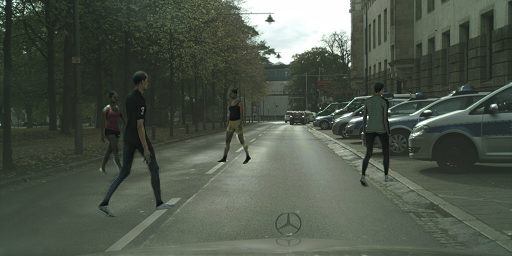}
\includegraphics[width=1\textwidth]{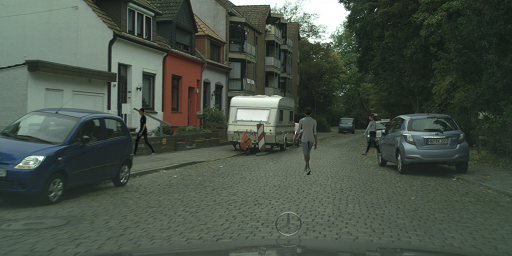}
\caption*{Adaptation}
\end{minipage}
\caption{Examples of domain adaptation by multi-discriminator architecture from augmented to Cityscapes}
\label{fig:adaptation}
\end{figure}

For evaluation of the translated results we conduct several experiments and assess our results both qualitatively and quantitatively. Some examples of the augmentation and adaptation are shown in figure~\ref{fig:adaptation}. In order to visualize the effects introduced by our adaptation approach we show both augmented and translated image together with manually picked crops where those effects are characteristically prominent. First, we want to draw attention to the fact that proposed multi-discriminator architecture effectively helps to solve the vanishing objects problem which came from discrepancy between distributions of augmented and original data. Provided samples reveal no semantic inconsistencies between synthetic and translated samples and in-painted objects are kept in place. Another characteristic of the translated images which is worth accounting is the actual appearance of the rendered objects. In the translated images rendered objects follow the color scheme of the entire target dataset. Finally, the feature which has been introduced in the transfered images is lighting  effects learned from the scenes. This could be observed on the magnified segments, showing applied light spots and smoothed edges.

In order to perform quantitative evaluation of the results we asses our generated data on two computer vision tasks: semantic and instance segmentation. For this purpose we utilize the state-of-the-art computer vision algorithms such as Deeplabv3 \cite{Chen17} and Mask-RCNN \cite{He2017}. 

\subsection{Instance Segmentation}
We evaluate image transfer quality on the task of instance segmentation reporting a standard COCO \textit{average precision(AP)} metric. We deploy one of the top performing detection models called Mask-RCNN \cite{He2017} which is pre-trained on COCO dataset and fine-tuned on our \textit{augmented cityscapes} dataset and evaluate on 500 images of Cityscapes \textit{val}. We follow the experimental setup of the original work \cite{He2017} and report the results for instance segmentation in table~\ref{tab:maskrcnn}.

\subsection{Semantic Segmentation}
Effectiveness of the generated data for semantic segmentation task was assessed with Deeplabv3 model. Similar to the preceding instance segmentation experiment we train baseline method on both Cityscapes (train) and our generated dataset and evaluate on Cityscapes (val). In both cases we down-sample the images to 1024$\times$512 pixel as a preprocessing step.
Deeplabv3 \cite{Chen17} uses \textit{xception65} backbone. It was trained for 90K iteration with batch 16 on 513$\times$512 random crops. Learning rate in this case remains 0.007. For the best performing snapshots \textit{meanIoU} metric is reported in table~\ref{tab:semseg}.\\
In both evaluation experiments our generated data shows the similar performance with regard to overall metrics such as \textit{meanIoU} and $AP_{avg}$ but reveals slight performance improving for our class of interest - pedestrian. Improvement for $AP_{pedestrian}$ in Mask-RCNN experiments is more than 2.5 percent. And improvement for Deeplab on pedestrian class is rather moderate (less than 0.5), at the same time pixel accuracy and meanIoU metric over all 19 classes shows almost no performance decrease (0.3 percent).

\begin{table}[!t]
\renewcommand{\arraystretch}{1.5}
\resizebox{\columnwidth}{!}{
\begin{tabular}{l|cc|cc}
\hline
\rotatebox[origin=c]{0}{ Method }
& \rotatebox[origin=c]{0}{ Accuracy }
& \rotatebox[origin=c]{0}{ mean IoU }
& \rotatebox[origin=c]{0}{ person }
\\
\hline

CS		& 95.6 & 75.6 & 77.1 \\
Ours	& 95.3 & 75.3 & 77.3 \\



\hline
\end{tabular}}
\caption{meanIoU values for semantic segmentation prediction by Deeplabv3 trained on Cityscapes and our dataset.}
\label{tab:semseg}
\end{table}


\begin{figure}[!h]

\begin{minipage}{.495\columnwidth}
\includegraphics[width=1\textwidth]{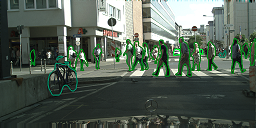}
\includegraphics[width=1\textwidth]{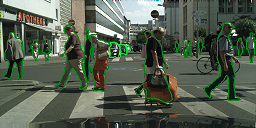}
\includegraphics[width=1\textwidth]{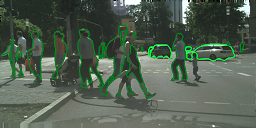}
\includegraphics[width=1\textwidth]{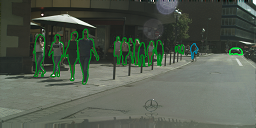}
\caption*{Instance segmentation}
\end{minipage}
\begin{minipage}{.495\columnwidth}
\includegraphics[width=1\textwidth]{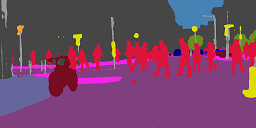}
\includegraphics[width=1\textwidth]{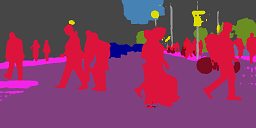}
\includegraphics[width=1\textwidth]{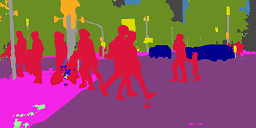}
\includegraphics[width=1\textwidth]{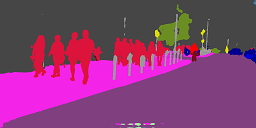}
\caption*{Semantic segmentation}
\end{minipage}
\caption{Results by Mask-RCNN and Deeplabv3 trained on our dataset.}
\label{fig:maskrcnn}
\end{figure}

\section{Conclusion}
In this paper we presented a pipeline for data generation which consists of pedestrian augmentation part and dataset appearance learning part based on novel class specific multi-discriminator architecture. During experiments it was shown that proposed pipeline can generate semantically and geometrically consistent training images with target dataset optic which helps to bridge domain gap between augmented and real data.

\newpage
{
\bibliographystyle{ieee}
\bibliography{literature}
}

\end{document}